\documentclass[11pt]{article}

\usepackage{latexsym,mathrsfs}
\usepackage{amsmath,amssymb}
\usepackage{amsthm,enumerate,verbatim}
\usepackage{amsfonts}
\usepackage{graphicx}
\usepackage{algorithm}
\usepackage{algorithmic}
\usepackage{url}

\newtheorem*{definition*}{Definition}

\newtheorem{remark}{Remark}

\newtheorem*{problem*}{Problem}

\numberwithin{equation}{section}
\numberwithin{table}{section}
\numberwithin{figure}{section}

\title{Neither Global Nor Local: A Hierarchical Robust Subspace Clustering For Image Data} 
\date{}

\author{Maryam Abdolali\thanks{Department of Computer Engineering and Information Technology, Amirkabir University of Technology, Tehran, Iran.
Email addresses: mabdolali@aut.ac.ir (Maryam Abdolali), rahmati@aut.ac.ir (Mohammad Rahmati)} 
 \and Mohammad Rahmati$^*$
}

\begin{document}

\maketitle

\begin{abstract}

In this paper, we consider the problem of subspace clustering in presence of contiguous noise, occlusion and disguise. We argue that self-expressive representation of data in current state-of-the-art approaches is severely sensitive to occlusions and complex real-world noises. To alleviate this problem, we propose a hierarchical framework that brings robustness of local patches-based representations and discriminant property of global representations together. This approach consists of 1) a top-down stage, in which the input data is subject to repeated division to smaller patches and 2) a bottom-up stage, in which the low rank embedding of local patches in field of view of a corresponding patch in upper level are merged on a Grassmann manifold. This summarized information  provides two key information for the corresponding patch on the upper level: cannot-links and recommended-links. This information is employed for computing a self-expressive representation of each patch at upper levels using a weighted sparse group lasso optimization problem. Numerical results on several real datasets confirm the efficiency of our approach.
\end{abstract}


\section{Introduction}

Detecting low-dimensional structures in high-dimensional data is essential to many applications in different areas including computer vision and image processing. In many of these applications, data often have fewer degree of freedom than the original high ambient dimension. This observation led to development of several classical approaches to find the low-dimensional representation for the data~\cite{cayton2005algorithms,hegde2008random,carreira1997review}. Huge family of these approaches, including the well-known Principal Component Analysis (PCA) method~\cite{jolliffe1986principal}, assumes that the data lie in a \emph{single} low-dimensional subspace/manifold. However, in many applications, data points are distributed around several subspaces and should better be represented by \emph{multipe} low-dimensional subspaces. Modeling data with collection of subspaces led to a more general problem that is often referred to as \emph{Subspace Clustering}~\cite{vidal2011subspace}: 

Definition 1 (Subspace Clustering): Let $X = {\{x_i \in \mathbb{R}^{D}\}}_{i=1}^N$ be a collection of data points from $n$ unknown subspaces $S_1, S_2, \cdots, S_n$  with intrinsic dimensions $d_1, d_2, \cdots, d_n$ (where $d_i << D$). The goal of subspace clustering is to segment the data points according to their underlying subspaces and to identify parameters of each subspace.

This problem has received considerable attention over the past decade and many attempts have been made to address it~\cite{kanatani2001motion,vidal2005generalized,tseng2000nearest,ho2003clustering,ma2008estimation,tipping1999mixtures,yan2006general}. Among them, a family of approaches based on recent developments in low-rank and sparse representation have achieved state-of-the-art results~\cite{elhamifar2009sparse,elhamifar2013sparse,liu2010robust,liu2013robust}. These algorithms have two major key steps: 1) Constructing an affinity matrix (neighborhood graph) using self-expressiveness property of data~\cite{elhamifar2009sparse,von2007tutorial}, which assumes that each data point can be written as linear combination of other data points, and 2) Obtaining the data clusters by applying spectral clustering on the affinity matrix. In particular, the affinity matrix in these algorithms is constructed by optimizing the following problem:
\begin{align} 
    \min_{C \in \mathbb{R}^{N \times N}, E \in \mathbb{R}^{D \times N}} & \; f(C) + \lambda g(E) \label{SSCLRRLSR} 
    \\  \text{ such that }  & \; X = XC + E \text{ and } 
    C_{i,i} = 0 \text{ for all } i, \nonumber
\end{align} 
where $X \in \mathbb{R}^{D \times N}$  is the data matrix (with $D$-dimensional data points as columns and $N$ the number of data points),  $C \in \mathbb{R}^{N \times N}$ is the coefficient matrix and $E \in \mathbb{R}^{D \times N}$ is the error matrix. $f(.)$ and $g(.)$ denote the regularizations on the matrices $C$ and $E$ respectively. The three main representative methods among these algorithms, namely SSC (Sparse Subspace Clustering)~\cite{elhamifar2013sparse}, LRR (Low Rank Representation)~\cite{liu2013robust} and LSR (Least Square Regression)~\cite{lu2012robust} differ in the function of $f(C)$. In particular, $f(C)$ is $\ell_1$ norm for SSC which prioritizes sparse solutions, nuclear norm (sum of singular values of input matrix) for LRR which prioritizes low rank solutions and Frobinuis norm for LSR. Once the problem in (\ref{SSCLRRLSR}) is optimized, the affinity matrix is obtained by symmetrizing the coefficient matrix $C$ via $|C| + |C^T|$. The clusters are next calculated by applying spectral clustering on the affinity matrix. 

Even though advances in sparse and low-rank representation literature contributed significantly to the development of elegant subspace clustering algorithms, these algorithms are based on \emph{global} self-expressive representation of data in which the \emph{entire} data point is expressed as linear combination of other data points. However, clustering based on global data representation can be easily affected by occlusions and severely corrupted noisy blocks~\cite{abdolali2019robust} and the conventional $\ell_1$ or Forbinuis norms which are often utilized to regularize the error matrix are incapable to model the complexity of real-world data corruptions~\cite{luo2016tree}. On the other hand, block-wise (local) representations tend to be more robust to occlusions and contiguous noises but combining the clustering results of the local patches by a trivial majority voting scheme might drastically be affected by non-informative patches.

 In order to overcome this major shortcomming, we propose an efficient hierarchical framework that combines the advantages of \emph{local} representation with global self-expressive representation of data. Specifically, in our approach, each data point is divided into a collection of blocks/patches and the local connectivities of these patches are summarized using the corresponding low-dimensional embeddings on a Grassmann manifold~\cite{turaga2008statistical}. This local connectivity information is then used to guide the global self expressive representation by providing prior knowledge on 1) which connections should be avoided using weighted sparse regularization and 2) which connections are recommended by the local representations using group sparse regularization~\cite{friedman2010note}.

 The main contribution of this paper is to  propose a novel framework, dubbed as LG-SSC (\emph{Locally-Guided SSC}), to bridge the gap between discriminative global representation and robust local alternative in order to achieve a robust discriminant representation for subspace clustering. The proposed framework includes of two key steps: 1) combining diverse characteristics of local patches using a hierarchical low-dimensional analysis on Grassmann manifolds and 2) computing global sparse self-expressive representation of data using group lasso and weighted sparse regularizations. As we will show, our approach is more robust to occlusions, illumination effects and continuous block noises.

This paper is an extension of the work originally presented in~\cite{abdolali2019local}. Inspired by the significant increase in clustering accuracy using the self-expressive representation of local patches in our preliminary work, we further utilized this information in a different hierarchical approach. In particular, in this paper, the local information fusion in each level is fed back to guide the calculation of coefficient matrix in upper level in order to achieve a more robust representation. This approach which can be considered as \emph{coefficient calculation-level} fusion exhibits stronger robustness and clustering accuracy compared to the previous \emph{subspace estimation-level} fusion.

The rest of the paper is organized as follows: In Section~\ref{relwork}, the related works are briefly reviewed. After presenting the motivation behind our proposed approach in Section~\ref{motivation}, the detailed explanation of the two suggested major steps is elaborated in Section~\ref{proposed}. We evaluate the performance of proposed approach  in Section~\ref{experiments} and finally conclude our work in Section~\ref{conclusion}.
\section{Related Works} \label{relwork}
Self-expressive based subspace clustering methods enjoy broad theoretical guarantees for data that is drawn perfectly from a collection of low-dimensional subspaces~\cite{soltanolkotabi2012geometric,elhamifar2013sparse} and recently these foundations have been extended to noisy cases as well~\cite{wang2016noisy}. 
 An intuitive simple way for modeling errors was first proposed in ~\cite{elhamifar2009sparse} where self-expressiveness term was relaxed to $X = XC + E$. The elements of error matrix $E$ were generally modeled by independent Laplacian or Gaussian distributions using $\ell_1$ or Frobinuis norms. Even though this has the advantage of simplicity and maintaining the convexity of the optimization problem, in practice error matrices are generated by more complicated variations and specifically the independence assumption between error elements in these models is too restrictive. 

To further improve robustness, other distributions were used to model the noise, including Cauchy~\cite{li2018robust} and Mixture of Gaussian~\cite{yao2018robust}. In ~\cite{vidal2014low} they consider the errors directly in the original space by proposing a non-convex formulation in which they assumed that the corrupted data are generated by linear combination of error matrix and clean data matrix and the self-expressiveness holds for the clean data. Even though these approaches show some improvements in dealing with practical noise, they still assume that the elements of the error matrix are independent. However, the contiguous error caused by occlusion, illumination effects and disguise does not follow this assumption.  

There are several approaches that tried to improve robustness in the representation space instead of the input space. Lu et al~\cite{lu2013correlation} utilized the relatively new Trace Lasso norm to regularize the coefficient matrix in order to improve the grouping effect. In~\cite{qiu2015learning}, an iterative approach was proposed in which a linear transformation was learned such that low-rank structures for data from the same subspaces were restored and a maximally separated structure for data from different subspaces was obtained. In ~\cite{li2017structured}, a unified optimization framework for learning both the coefficient matrix and the segmentation was presented. In this iterative approach, at each iteration, a segmentation matrix was constructed to help re-weighting the representation matrix in order to avoid certain connections. Lu at al~\cite{lu2019subspace} proposed a block diagonal induced regularizer to explicitly enforce the Laplacian matrix corresponding to the coefficient matrix to be block diagonal. This enforces the neighborhood graph to contain exactly $n$ connected components and might reduce the number of wrong connections. 

Although the above approaches could enhance the robustness in some cases, an important shortcoming is that they are based on global representation and hence can be easily affected by severely corrupted regions in data points. Patch-based image representations were often used to increase robustness to continuous noises in sparse representation literature for classification~\cite{zhu2012multi,tan2005recognizing,lai2016classwise}. Multi-scale patch based regularization using nuclear norm for modeling the error has shown effectiveness for face classification~\cite{luo2016tree} and single low-rank subspace estimation~\cite{abdolali2017multiscale}. In our previous work, MG-SSC (Multi-Graph based SSC) ~\cite{abdolali2019local}, a multi-layer graph was constructed by dividing each data into patches in different sizes and computing the sparse affinity matrix corresponding to each collection of patches using SSC. A summarized low-dimensional representation of this multi-layer graph is computed using a weighted subspace analysis of individual graphs on a Grassmann manifold. In this paper, the advantages of global and local (patch-based) representations are brought together for the problem of subspace clustering in a novel different framework. In the following sections, the details of the proposed approach are discussed thoroughly.

\section{The importance of local representations in robust subspace clustering} \label{motivation}
In many practical subspace clustering cases, the data could be partially occluded or corrupted. Even in these cases, detecting the low-dimensional structures may be still possible using the redundancy that is often present in high-dimensional data. However, corrupted samples might affect the neighborhood graph severely such that they lead to a completely wrong understanding of data structure. 

\begin{figure}[h!]
\begin{center}
\includegraphics[width=10cm]{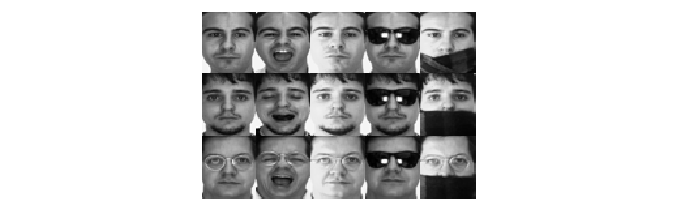} 
\caption{Few samples corresponding to three individuals of AR data base}
\label{fig1}
\end{center}
\end{figure}

We illustrate this concept with a real world example. We consider facial images of three subjects from AR database~\cite{martinez1998ar} (more on this database in the experiment section). Few samples from the selected images are shown in Figure \ref{fig1}. As it can be seen several varieties of corruptions are present in these images, including disguise using sun glasses, scarves and illumination variations. It is usually assumed that facial images of a subject under different lighting conditions can be approximated by a nine dimensional linear subspace~\cite{basri2003lambertian} and hence, a collection of facial images of several subjects lie on a union of nine dimensional subspaces~\cite{elhamifar2009sparse}. 

\begin{figure*}[htb]
\begin{minipage}[b]{0.3\linewidth}
  \centering
  \centerline{\includegraphics[width=6cm]{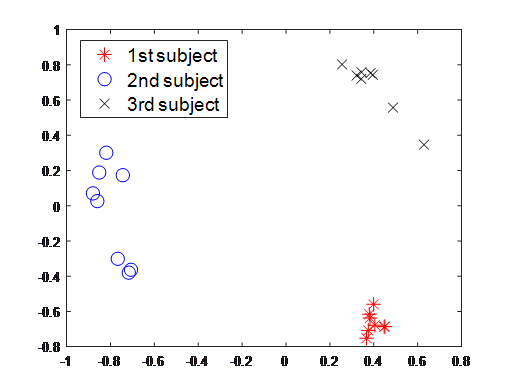}}
  \centerline{(a)}\medskip
\end{minipage}
\hfill
\begin{minipage}[b]{0.3\linewidth}
  \centering
  \centerline{\includegraphics[width=6cm]{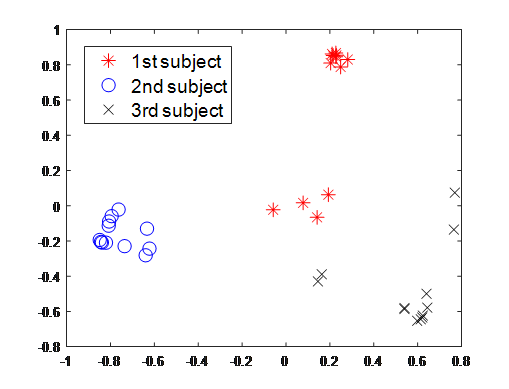}}
  \centerline{(b)}\medskip
\end{minipage}
\hfill
\begin{minipage}[b]{0.3\linewidth}
  \centering
  \centerline{\includegraphics[width=6cm]{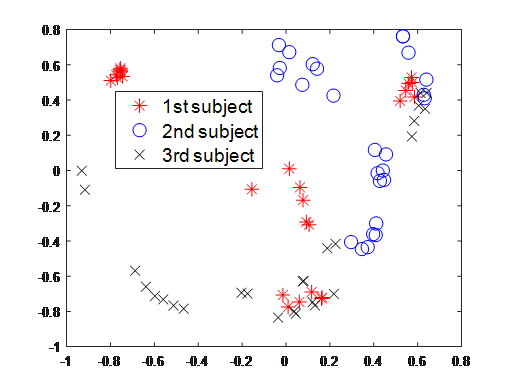}}
  \centerline{(c)}\medskip
\end{minipage}
\caption{SSC based embedded vectors of samples (a) without any illumination variations or occlusions (b) with illumination variations but without occlusions (c) with both illumination variations and occlusions.}
\label{effects_3}
\end{figure*}

We select three subsets $\{X^{(i)}\}_{i=1}^3$ from these images: $X^{(1)}$ which includes images with neither illumination effect nor disguise, $X^{(2)}$ which includes images with no disguise but under different illumination effects and $X^{(3)}$ which includes the whole images under both variations. The SSC algorithm is applied on these subsets of images separately. The normalized spectral embedding of the three corresponding coefficient matrices $\{C^{(i)}\}_{i=1}^3$ are plotted in Figure \ref{effects_3}. In particular, for each coefficient matrix $C^{(i)}$, the corresponding Laplacian matrix $L^{(i)}$ is calculated as $L^{(i)} = I - {D^{(i)}}^{-\frac{1}{2}}C^{(i)}{D^{(i)}}^{-\frac{1}{2}}$ where $D^{(i)}$ is the diagonal matrix with its $(k,k)$th entry defined as $D^{(i)}_{kk} = \sum_{j} C^{(i)}_{kj}$. The normalized eigenvectors corresponding to the second and third smallest eigenvalues of the Laplacian matrix provide a 2-dimensional representation for the data affinity. The representations for three individuals are plotted in three different colors (red, blue and black). By comparing the three embedded data in Figure~\ref{effects_3} (a) - (c), it can be seen that illumination and especially occlusion can affect the \emph{global} self-expressive coefficients. Especially the occlusion (sun glasses and scarves) makes the detection of the three subspaces impossible.

\begin{figure*}[htb]
\begin{minipage}[b]{0.5\linewidth}
  \centering
  \centerline{\includegraphics[width=9cm]{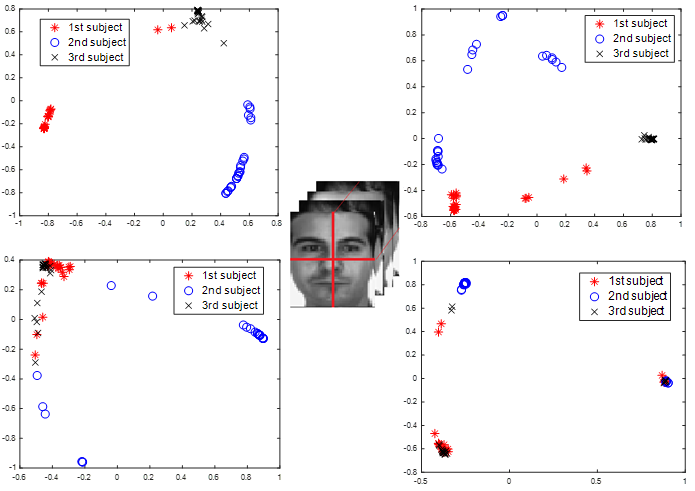}}
  \centerline{(a)}\medskip
\end{minipage}
\hfill
\begin{minipage}[b]{0.2\linewidth}
  \centering
  \centerline{\includegraphics[width=4cm]{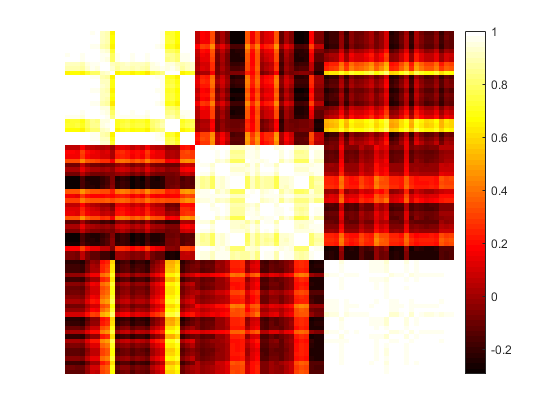}}
 \vspace{1.5cm}
  \centerline{(b)}\medskip
\end{minipage}
\hfill
\begin{minipage}[b]{0.2\linewidth}
  \centering
  \centerline{\includegraphics[width=5cm]{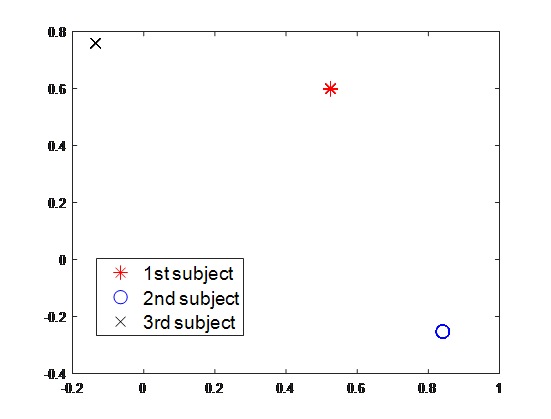}}
 \vspace{1.5cm}
  \centerline{(c)}\medskip
\end{minipage}
\caption{The effect of local representations. (a) The embedded vectors of samples corresponding to 4 patches, (b) the obtained summary representation (spectral kernel) and (c) the final embedded vectors of samples  using proposed framework }
\label{effects}
\end{figure*}

In contrast to global representations, patch-based representations can be more robust to severely corrupted regions. To illustrate this, we divide each facial image in $X^{(3)}$ (which contains all images of the three selected individuals) to 4 patches and plot the spectral embedding of the 4 patches right next to the corresponding patch  in Figure~\ref{effects}.\footnote{The spectrum of each patch is obtained using the proposed algorithm by dividing it into 4 more patches. We explain this in the next section.} As it can be noticed, each local representation provides some information on the connectivities within data points, however, in almost none of them, the 3 clusters are perfectly separated. Moreover, with no prior knowledge we cannot identify the best representation among these 4 representations. Our proposed framework, based on multi-layer graph fusion, obtains a (spectral kernel) summary representation from these local representations (which is shown in Figure~\ref{effects} (b) for complicity). This summary representation is a $N \times N$ matrix with values between -1 and 1, whose entries quantify the possibility of two corresponding data points belonging to the same subspace according to the local representations. The values which are closer to 1 indicate the higher possibility. This information is used to improve the quality of global representation (details are presented in the next section). The final global low-dimensional embedding corresponding to our proposed LG-SSC is plotted in Figure~\ref{effects} (c) and it can be already seen that not only the three clusters are perfectly separated but also all the samples within each cluster have the exact same embedding in this case. This suggests that our proposed framework can improve within clusters connectivities along with overall clustering accuracy.   
\section {Proposed Framework: Locally-Guided SSC} \label{proposed}
The proposed hierarchical framework, dubbed as LG-SSC, consists of a top-down stage and a bottom-up stage. LG-SSC has two major ingredients in each level of this hierarchy: 1) Dividing the data into local patches (in top-down stage) and summarizing the connectivity information between them (in bottom-up stage) and 2) guiding the detection of low-dimensional structures using this information (in bottom-up stage). In this section, details of these two key parts are presented.
\subsection{Division and local information summarization}
\begin{figure}[h!]
\begin{center}
\includegraphics[width=9cm]{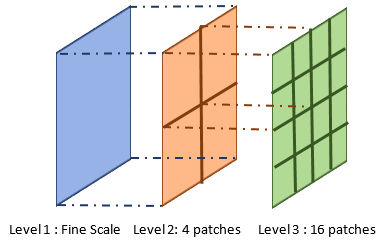} 
\caption{Dividing sample $x_i$ (in 2d image form) using the hierarchical structure corresponding to $s=3$ (3 levels) and $p=4$ (4 patches). The field of view of two patches in level 1 and level 2 are illustrated for clarity as well.} 
\label{3levels}
\end{center}
\end{figure}

Given the data matrix $X = [x_1, x_2,...,x_N]$, we construct a hierarchical structure composed of $s$ levels. At the top of this structure is the given data matrix. In the next level, the image data is divided into $p$ patches (typically 4 or 9). The division is carried out for each image data $x_i$ in its (2d) matrix form. This process is repeated for each patch in the next level. The created $p$ patches indicate the \emph{field of view} of the corresponding patch from upper level. This procedure is shown in Figure~\ref{3levels} for $s=3$ and $p=4$ for one image data. Let the whole gallery patches at level $i$ be represented by $X^{(i)}=\{X^{(i)}_1, ..., X^{(i)}_T\}$ where $T=p^{(i-1)}$ is the number of patches that are generated in the level of $i$ and $X^{(i)}_j$ ($j=1,...,T$) is the collection of local patches (at the $i$th level) with same coordination (position) from all images. Clearly the patches in the lower levels are smaller in size and generally contain less discriminative information about the whole image data. However, they are more robust to corruptions.
Suppose that in the $i$th level, the collection of patches $X^{(i)}_j$ (where $j \in \{1,...,p^{i-1}\}$) are further divided to $p$ patches. Hence, at the $(i+1)$th level, $p$ coefficient matrices can be obtained corresponding to the set of $p$ collected patches. We denote these patches and their corresponding coefficient representations by $\{X^{(i+1)}_{\ell_j^k}\}_{k=1}^p = \{X^{(i+1)}_{\ell_j^1},...,X^{(i+1)}_{\ell_j^p}\}$ and $\{C^{(i+1)}_{\ell_j^k}\}_{k=1}^p = \{C^{(i+1)}_{\ell_j^1},...,C^{(i+1)}_{\ell_j^p}\}$  respectively. $\ell_j^k$ ($k=1,...,p$) indicates the indices of collection of patches that are generated by dividing the $j$th patch in the upper level. These patches are in the field of view of the collection of patches in $X^{(i)}_j$. The set  $\{C^{(i+1)}_{\ell_j^k}\}_{k=1}^p$ contains $p$ different types of relationships between the data points. With no prior knowledge, it is highly non-trivial to choose the best representation among them. In fact, different coefficient matrices can be interpreted as different \emph{views} from the larger patches at the finer level where each coefficient matrix corresponds to an affinity matrix of a graph. Hence, in order to combine the information of the different coefficient matrices, we use a multi-layer graph fusion approach~\cite{dong2014clustering} by transforming the information of each representation into a subspace on the Grassmann manifold. 

The utilized graph fusion approach~\cite{dong2014clustering} is briefly presented as follows: Given $p$ different graph affinity matrices with common vertex set representing the data points, the goal is to merge the information provided by different individual graphs. To summarize the intrinsic relationships between the data points (the vertices of the graph), the problem can be mapped to the problem of finding a low-dimensional representation such that it preserves information of each affinity matrix in a meaningful way.\\
As mentioned in Section 2, for each graph affinity matrix ($A^i_{\ell_j^k}$), there is a corresponding normalized spectral embedding  ($U^i_{\ell_j^k}$) which can be considered as the low-dimensional representation of the affinity matrix. This spectral embedding can be obtained by simply calculating the $n$ eigenvectors corresponding to $n$ smallest eigenvalues of the Laplacian matrix (from the affinity matrix). Using the collection of subspace representations of the $p$ available affinity matrices, one can naturally define a summary representation of multiple affinity matrices by optimizing the following problem~\cite{dong2014clustering}:
\begin{equation}
\begin{split}
    V^{(i-1)}_j=&\arg\min_{V \in \mathbb{R}^{N \times n}}  \sum_{k=1}^p tr(V^T L^i_{\ell_j^k}V) - \alpha \sum_{k=1}^p tr(VV^T U^i_{\ell_j^k}{U^i_{\ell_j^k}}^T) \label{MG-SSC} 
    \\ & \text{ such that } \; V^TV=I, \nonumber
\end{split}
\end{equation}
where $\alpha$ is the regularization parameter. The first term in the above optimization problem, attempts to find a summary representation $V$ from the collection of subspace representations $\{U^i_{\ell_j^k}\}^p_{k=1}$ such that the connectivity information of each individual affinity matrix is preserved. The second term is the sum of squared projection distances between $V$ and individual subspaces  $\{U^i_{\ell_j^k}\}^p_{k=1}$ on a Grassmann manifold. This term enforces the summary representation to be close to other subspace representations on the Grassmann manifold.
This optimization problem has a closed form solution based on Rayleigh-Ritz theorem~\cite{von2007tutorial}. The summary representation $V$ can be obtained by calculating the smallest $n$ eigenvectors of the following matrix:
\begin{align}
(L^{(i)}_j)_{summary}=\sum_{k=1}^p  L^i_{\ell_j^k} - \alpha \sum_{k=1}^p U^i_{\ell_j^k}{U^i_{\ell_j^k}}^T.
\end{align}
Each row of the matrix $V^{(i-1)}_j$ is normalized to have unit $\ell_2$ norm. As we shall see, the subspace representation of the $p$ local patches in the $i$th level can be used to calculate the coefficient matrix of the corresponding pacth in the higher level ($i-1$) more efficiently.
\subsection{From local summarization to global representations}
\begin{figure*}[h!]
\begin{center}
\includegraphics[width=13cm]{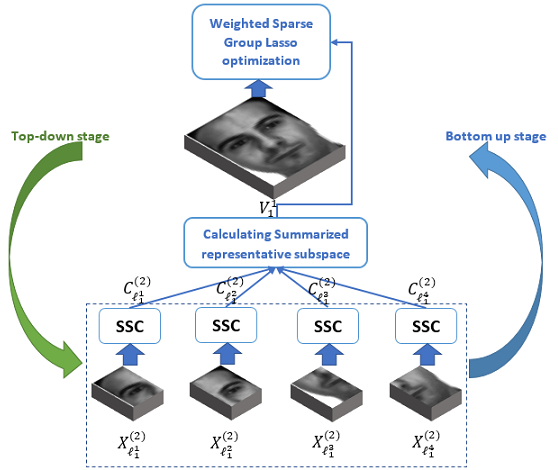} 
\caption[justification=centering]{Illustration of steps of LG-SSC for $p=4$ and $s=2$. In the top-down stage, each image is divided into $p$ patches to obtain $\{X^{(2)}_{\ell_1^k}\}_{k=1}^4$ in the second level. In bottom-up stage, SSC is applied on each collection of patches and the obtained coefficient matrices are merged using corresponding subspace analysis on Grassmann manifold. The summarized information is employed to guide the calculation of final global coefficient matrix in the fine scale. } 
\label{overview}
\end{center}
\end{figure*}

Following dividing the image data into local patches in different scales, and in order to obtain the global coefficient matrix, a bottom-up process should be carried out. For a clear explanation, suppose we have only two levels ($s=2$) and each data point (an image) is divided into 4 patches ($p=4$).
The overview of LG-SSC for $p=4$ and $s=2$ is illustrated in Figure~\ref{overview}.
In particular, we start from the coarsest level (second level here) and the regular SSC algorithm is applied on the collection of patches in this level. Hence, 4 sets of affinity matrices $\{C^2_{\ell_1^k}\}^4_{k=1}$ are obtained. 

Using the methodology in previous subsection, a summary representation ($V^1_1 \in \mathbb{R}^{N \times N}$) from the 4 local representations can be calculated efficiently. We define the spectral kernel of the summary representation in the embedded space as:
\begin{align}
   K^1_1 =(V^1_1)(V^1_1)^T, \nonumber
\end{align}
The matrix $K^1_1 \in [-1:1]^{N \times N}$ provides information on the similarity between samples in the embedded space. The entries with low values indicate the connectivites which local representations highly agree that should be avoided. Therefore, we can indicate the \emph{cannot-links} by adding an extra penalty on coefficient matrix when the values of $K^1_1$ are low. We define the matrix $\Theta^1_1 \in \mathbb{R}^{N \times N}$ as:
\begin{align}
    \Theta^1_1 = (1-K^1_1), \nonumber
\end{align}
The entries of the matrix $\Theta^1_1$ have high values when the corresponding entries in $K^1_1$ have low values. Using this matrix, we obtain the coefficient matrix $C^1_1$ in the fine level (global representation in this example) by optimizing the following problem:
\begin{align}
    C^1_1 = \arg &\min_{C \in \mathbb{R}^{N \times N}} ||C||_1 + \lambda_1||\Theta_1^1 \odot C||_1 + \frac{\mu}{2}||X - XC||^2_F \label{mid}
    \\ & \text{ such that } \; diag(C)=0, \nonumber
\end{align}
where $\lambda_1$ and $\mu$ are regularization parameters and $\odot$ denotes the element-wise (Hadamard) production. This is in fact a weighted lasso optimization problem, where we are adding an extra penalty on the entries of the matrix $C$ when the corresponding entries in $\Theta_1^1$ are high. This can be validated by rewriting the first two terms as:
\begin{align}
||C||_1 + \lambda_1||\Theta_1^1 \odot C||_1 = \sum_{i,j} |C_{ij}| (1+\lambda_1 (\Theta_1^1)_{ij})
\end{align}
where $(\cdot)_{ij}$ denotes the (i,j)the entry of the input matrix. 
\begin{remark}
The elements of the matrix $K_j^i$ (here $i=j=1$) can be interpreted as a quantified measure of the possibility that the two corresponding samples belonging to the same subspace according to the obtained summary representation. In particular, the closer the values of these elements are to one, the higher is the possibility of the two samples belonging to the same subspace. Hence, we can drop the additional penalty on entries with high values, specifically the ones higher than a predefined threshold. In practice, this threshold can be set to any value in the range $[0.7-0.9]$ as the proposed framework is not sensitive to this value.
\end{remark}

However, the summary representation $V^1_1$ not only indicates the \emph{"cannot-links"} constraints but also contains information about which connections are \emph{recommended} by local views as well. By applying k-means algorithm on the normalized rows of $V^1_1$, an initial grouping of data points based on the summary of local representation is achieved. This grouping information denotes the connectivites that the local representations recommend. Let $G$ be the set that contains the indices of data points in each cluster according to $V^1_1$. This information is added to the problem (\ref{mid}) as following:
\begin{equation}
\begin{split}
    C^1_1 = \arg\min_{C \in \mathbb{R}^{N \times N}} &  ||C||_1 + \lambda_1||\Theta_1^1 \odot C||_1 \\
 &+ \lambda_2 \sum_{j=1}^N \sum_{g \in G} ||(C_{:j})_g||_2 +  \frac{\mu}{2}||X - XC||^2_F \label{final}
    \\  & \text{ such that } \; diag(C)=0, 
\end{split}
\end{equation}
where the third term is the group lasso regularization~\cite{friedman2010note}. Here $(\cdot)_{:j}$ denotes the $j$th column of the input matrix and $(\cdot)_g$ indicates the input matrix's rows within the set $g$. Group lasso~\cite{friedman2010note} is the generalization of standard lasso in which the entries of coefficient matrix corresponding to the indicated groups are either included or excluded from the regression model together. Using this regularization, we encourage the samples to get connected to the samples within the recommended groups (improving connectivity) and simultaneously to get disconnected from the groups that are recommended to be avoided. 

This procedure can be easily extended to cases with higher number of levels ($s>2$). In general, at the bottom level of this hierarchical structure, the typical SSC is applied on the set of patches $\{X_j^s\}^{p^{s-1}}_{j=1}$. In the upper levels, the explained method (with two key steps) is applied on the gallery of patches individually and this procedure is continued until the root level which outputs the final \emph{global} coefficient representation and the corresponding segmentation result.

The proposed framework not only computes the segmentation based on the global discriminant representation of the data but also takes advantage of robust local knowledge in each level of the hierarchy, leading to a robust discriminant representation.

\subsection{Optimization}
In this section, we introduce the optimization of the proposed convex problem in (\ref{final}) using Alternating Direction Method of Multipliers (ADMM)~\cite{boyd2011distributed}. First, we introduce auxiliary matrix $Z \in \mathbb{R}^{N \times N}$ and consider the following problem (The superscripts and subscripts in the optimization problem are dropped for ease of reading):
\begin{equation}
\begin{split}
     \min_{C \in \mathbb{R}^{N \times N}, Z \in \mathbb{R}^{N \times N}} &  ||C||_1 + \lambda_1||\Theta \odot C||_1 
	+ \lambda_2 \sum_{j=1}^N \sum_{g \in G} ||(C_{:j})_g||_2 +  \frac{\mu}{2}||X - XZ||^2_F 
    \\  & \text{ such that } \; Z = C - diag(C). \label{ADMM}
\end{split}
\end{equation}
Next, using penalty parameter $\beta>0$ and matrix $\Delta \in \mathbb{R}^{N \times N}$ of Lagrange multipliers for the equality constraint, we get the following problem:
\begin{equation}
\begin{split}
     \mathcal{L}(C,Z,\Delta) = ||C||_1 + \lambda_1||\Theta \odot C||_1 + \lambda_2 \sum_{j=1}^N \sum_{g \in G} ||(C_{:j})_g||_2 
	\\+  \frac{\mu}{2}||X - XZ||^2_F + \frac{\beta}{2}||Z - (C -diag(C))||_F^2 
	\\+ tr(\Delta^T(Z-C+diag(C))), \nonumber
\end{split}
\end{equation}
where $tr(.)$ is the trace operator. Based on ADMM, this problem is optimized using the following iterative procedure:

With an abuse of notation, let $(C^{(k)},Z^{(k)})$ be the optimization variables at iteration $k$ and $\Delta^{(k)}$ be the Lagrangian multiplier at the same iteration:
\begin{itemize}
\item Find $Z^{(k+1)}$, by minimizing $\mathcal{L}$ with respect to $Z$, while the rest of the variables are fixed by setting the derivative of $\mathcal{L}$ with respect to $Z$ to zero:
 \begin{align}
    (\mu X^TX+\beta I)Z^{(k+1)} = \mu X^TX + \beta C^{(k)} - \Delta^{(k)}. \nonumber
\end{align}
The matrix $Z^{(k+1)}$ is obtained by solving the above system of linear equations using approaches such as conjugate gradient method.
\item Find $C^{(k+1)}$, by minimizing $\mathcal{L}$ with respect to $C$, while the other variables are fixed. Note that we can rewrite this as:
\begin{equation}
\begin{split}
    \min_C ||W \odot C||_1 + \lambda_2 \sum_{j=1}^N \sum_{g \in G} ||(C_{:j})_g||_2 
	+ \frac{\beta}{2}||Z - C||_F^2 + tr(\Delta^T(Z-C)), \nonumber
\end{split}
\end{equation}
where $(i,j)$th entry of W is defined by:$W_{(ij)}=1+\lambda_1 \Theta_{ij}$ and we can simplify it more as:
\begin{equation}
\begin{split}
     \min_C ||W \odot C||_1 + \lambda_2 \sum_{j=1}^N \sum_{g \in G} ||(C_{:j})_g||_2 
	+ \frac{\beta}{2}||C - (Z+\frac{\Delta}{\beta})||_F^2. \nonumber
\end{split}
\end{equation}
The constraint of $diag(C) = 0$ can be considered by projecting the solution to the above problem on this constraint. This problem is group-separable, so we consider:
\begin{equation}
\begin{split}
     \min_{(C_{:j})_g} ||(W_{:j})_g \odot (C_{:j})_g||_1 + \lambda_2  ||(C_{:j})_g||_2 
	+ \frac{\beta}{2}||(C_{:j})_g - ((Z_{:j})_g+\frac{(\Delta_{:j})_g}{\beta})||_F^2. \nonumber
\end{split}
\end{equation}
By taking the (sub)gradient of the above problem, we have:
\begin{equation}
\begin{split}
	 (W_{:j})_g \odot  (S_{:j})_g &+ \lambda_2 \frac{(C_{:j})_g}{||(C_{:j})_g||_2} \\&+ \beta \left(C_{:j})_g - \left(Z_{:j})_g + \frac{(\Delta_{:j})_g}{\beta}\right)\right),\label{subg}\\
\end{split}
\end{equation}
where S is a matrix which is defined as following:
\begin{equation*}
\begin{split}
[{(S_{:j})_g}]_k = 
    \begin{cases}
    1 &  [{(C_{:j})_g}]_k < 0\\
    [-1,1] & [{(C_{:j})_g}]_k = 0\\
    -1 & [{(C_{:j})_g}]_k > 0
    \end{cases}
\end{split}
\end{equation*}
$[{(S_{:j})_g}]_k$ denotes $k$-th element of the vector $(S_{:j})_g$. By setting~(\ref{subg}) to zero:
\begin{equation*}
\begin{split}
	(C_{:j})_g +  \frac{\lambda_2}{\beta} \frac{(C_{:j})_g}{||(C_{:j})_g||_2}  &=\\( Z_{:j})_g + &\frac{(\Delta_{:j})_g}{\beta}-\frac{(W_{:j})_g \odot  (S_{:j})_g}{\beta}.
\end{split}
\end{equation*}

Hence the matrix C can be updated as:
\begin{equation}
\begin{split}
(J_{:j})_g & = \mathcal{T}_b\left((Z_{:j})_g+\frac{(\Delta_{:j})_g}{\beta}-\frac{(W_{:j})_g \odot (S_{:j})_g}{\beta},\frac{\lambda_2}{\beta}\right),\\ & \qquad\qquad\qquad \text{for }j \in [1,...,N] \quad \& g \in G \\
C^{(k+1)}& =J-diag(J) \label{mid}
\end{split}
 \end{equation}
where:
\begin{align}
\mathcal{T}_b(x,\rho)= \frac{x}{||x||_2}max\{0,||x||_2-\rho\}. \nonumber
\end{align}
Using this definition, (\ref{mid}) can be further summarized as:
\begin{equation}
\begin{split}
(J_{:j})_g & = \mathcal{T}_b(\mathcal{T}((Z_{:j})_g+\frac{(\Delta_{:j})_g}{\beta},\frac{(W_{:j})_g}{\beta}),\frac{\lambda_2}{\beta}),\\ & \qquad\qquad\qquad \text{for }j \in [1,...,N] \quad \& g \in G \\
C^{(k+1)}& =J-diag(J).
\end{split}
 \end{equation}
Here $\mathcal{T}$ is the shrinkage-thresholding operator and is defined as:
\begin{align}
\mathcal{T}(x,\rho )=(|x|-\rho)_+sign(x) \nonumber
\end{align}
\item Find $\Delta^{(k+1)}$ by gradient ascend:
 \begin{align}
    \Delta^{(k+1)} = \Delta^{(k)} + \beta (Z^{(k+1)} - C^{(k+1)}) \nonumber
\end{align}
\end{itemize}
These steps are repeated until convergence is met.  Note that the problem in (\ref{ADMM}) is convex and consists of two separable blocks of variables, hence, the solution obtained by ADMM is guaranteed to be optimal. Algorithm~\ref{algo1} summarizes the updates for the ADMM implementation.

\algsetup{indent=2em}
\begin{algorithm}[ht!]
\caption{Optimizing (\ref{ADMM}) via ADMM\label{algo1}}
\begin{algorithmic}[1] 
\REQUIRE 
$X \in \mathbb{R}^{D \times N}$, $G$ as the indices of data points in each cluster,
parameters $\lambda_1$, $\lambda_2$, $\mu$.
\ENSURE Coefficient matrix C. 
    \medskip  

\STATE Initialization: 
$k=0$, Initialize $C^{(0)}$, $Z^{(0)}$ and $\Delta^{(0)}$ to zero.

\WHILE{some convergence criterion is not met }

	\STATE  Update $Z^{(k+1)}$ by solving following system of linear equations:
\begin{align}
    (\mu X^TX+\beta I)Z^{(k+1)} = \mu X^TX + \beta C^{(k)} - \Delta^{(k)} \nonumber
\end{align}
	
	\STATE  Update $C^{(k+1)}$ as: 
\begin{equation}
\begin{split}
(J_{:j})_g & = \mathcal{T}_b(\mathcal{T}((Z^{(k+1)}_{:j})_g+\frac{(\Delta^{(k)}_{:j})_g}{\beta},\frac{(W_{:j})_g}{\beta}),\frac{\lambda_2}{\beta}),\\ & \qquad\qquad\qquad \text{for }j \in [1,...,N] \quad \& g \in G \\
C^{(k+1)}& =J-diag(J).\nonumber
\end{split}
 \end{equation}
	
	\STATE  Update $\Delta^{(k+1)}$ as:
	\begin{align}
		\Delta^{(k+1)} = \Delta^{(k)} + \beta (Z^{(k+1)} - C^{(k+1)}) \nonumber
	\end{align}
	
\ENDWHILE

\end{algorithmic}  
\end{algorithm} 

\section{Experiments} \label{experiments}
In this section, we demonstrate the effectiveness of LG-SSC in presence of illumination effects, occlusion and disguise using 3 publicly available databases for face and object clustering: Extended Yale B~\cite{georghiades2001few,lee2005acquiring}, AR database~\cite{martinez1998ar} and Coil-20~\cite{nene1996columbia}. The algorithm and all the experiments are implemented in Matlab and run on a computer with Intel Core i7-3770 CPU, 3.40 GHZ, 16 GB RAM. 

\textbf{Evaluation metrics:} To evaluate the quality of clustering, we use three well-known metrics, namely Accuracy (ACC), Normalized Mutual Information (NMI)~\cite{vinh2010information} and Adjusted Rand Index (ARI)~\cite{hubert1985comparing}. The accuracy of the clustering algorithm is calculated by following formula: 

\begin{align}
\text{ACC (\%)} = \frac{\text{\# of correctly classified points}}{\text{total \# of points}} \times 100. \nonumber
\end{align}
In NMI, the mutual information between segmentation result and ground-truth clusters is calculated and is then scaled between 0 and 1. ARI computes the Rand Index score and corrects it for chance. We multiply the values of NMI and ARI by 100 to have an easier comparison with ACC.

\textbf{Compared algorithms:} The performance of proposed LG-SSC algorithm is compared with 6 subspace clustering algorithms: SSC~\cite{elhamifar2013sparse}, LRR~\cite{liu2013robust}, EDSC~\cite{ji2014efficient}, $S^3C$~\cite{li2017structured}, LRSC~\cite{vidal2014low} and MG-SSC~\cite{abdolali2019local} (our preliminary work). The parameters of each algorithm are tuned for the best result and are reported in Table~\ref{tabpar} for each database.
\subsection{Extended Yale B face data set}
The Extended Yale B database~\cite{georghiades2001few,lee2005acquiring} contains 2,414 frontal-face images of 38 humans. There are 64 images, each of the size $192 \times 168$ pixels, per individual. The face images were captured under various lighting conditions. Similar to ~\cite{elhamifar2013sparse}, the images were downsampled to $48\times42$ pixels. For LG-SSC, we set $p=4$ and $s=2$. In order to study the effect of the number of clusters on the clustering performance, we implement 2 different settings of experiments: 1) We follow the setting utilized in ~\cite{elhamifar2013sparse}, which has been implicitly specified as the general setting for reporting the performance of subspace clustering algorithms on this database over the past years. In particular, for $n \in \{2,3,5,8,10\}$ clusters, the images of 38 subjects are divided into 4 groups of [1-10], [11-20], [21-30] and [31-38]. For $n \in \{2,3,5,8\}$ clusters, all choices of possible different trials for each group is considered and for $n=10$ subjects, only the first three groups are considered. The subspace clustering algorithms are applied on the corresponding subsets of images and the average ACC, NMI and ARI values over these subsets are reported in Table~\ref{tabyale}. The numbers indicated with * are taken from the corresponding papers. 2) For $n \in \{15, 20, 30, 38\}$, we simply select the first $n$ images of the database and apply the subspace clustering algorithms. The values of three metrics ACC, NMI and ARI for each subspace clustering algorithm is reported in Table~\ref{tabyale2}.

We observe that:
\begin{itemize}
\item LG-SSC and MG-SSC significantly outperform other approaches in all cases. Specifically, for $n \geq 15$, the accuracy of LG-SSC is more than 20\% higher than SSC which is the basic foundation of this approach. The results indicate the efficiency of hierarchical structure of LG-SSC in dealing with sever illumination effects.
\item MG-SSC slightly performs better than LG-SSC when the number of clusters is low. However, by increasing the number of clusters to 20, LG-SSC outperforms MG-SSC. This confirms that by gradually feeding summarized information of local patches in a hierarchical framework, the robustness can increase, especially in more challenging cases.
\item The performance of LG-SSC is quite stable with respect to the number of clusters.
\item The performance of SSC, LRR, $S^3C$ and LRSC decreases significantly as the number of clusters increases.
\item Sparse-based approaches, in particular SSC and $S^3C$, perform generally better compared to low-rank based approaches of LRR and LRSC.
\item EDSC benefits from a specific post-processing step which tends to be different from the other 6 approaches and this post-processing of affinity matrix plays a major role for the accuracy of clustering. We noted that without this post-processing step, the quality of the obtained coefficient matrix of EDSC is similar to LRR and LRSC. This makes sense as EDSC reguralizes the coefficient matrix using Frobinius norm which exhibits similar characteristics as nuclear norm in subspace clustering.
\end{itemize}

\begin{center}
\begin{table*}[!htbp]
\begin{center}
\caption{Parameters of the compared approaches.}
\label{tabpar}  
\small\addtolength{\tabcolsep}{-1pt}
\begin{tabular}{c||ccc}
\hline
	 \multicolumn{1}{c||}{Approach} & \multicolumn{3}{c}{Parameters} \\
	  & Extended Yale B & AR &Coil 20\\
\hline
  LGSSC & \begin{tabular}{@{}c@{}}$\alpha = 20$, $\lambda_1 = 1$, $\lambda_2 = 10$ \\ $p=4$, $s=2$\end{tabular} & \begin{tabular}{@{}c@{}}$\alpha = 100$, $\lambda_1 = 5$, $\lambda_2 = 10$\\ $p=4$, $s=3$\end{tabular}  &  \begin{tabular}{@{}c@{}}$\alpha = 20$, $\lambda_1 = 2$, $\lambda_2 = 10$\\ $p=4$, $s=2$\end{tabular}\\
MG-SSC & $\alpha = 20$, $p=4$, $s=3$ & $\alpha = 100$, $p=9$, $s=3$ & $\alpha = 20$, $p=4$, $s=2$\\
SSC & $\alpha = 20$ & $\alpha = 100$ & $\alpha =20$\\
LRR & $\lambda = 0.009$ & $\lambda = 0.095$ & $\lambda = 0.0092$\\
EDSC &  \begin{tabular}{@{}c@{}}$\lambda_1 = 0.06$, $\lambda_2 = 0.01$ \\  dim = 10, $\alpha = 4$\end{tabular} &   \begin{tabular}{@{}c@{}}$\lambda_1 = 0.06$, $\lambda_2 = 0.01$ \\  dim = 10, $\alpha = 4$\end{tabular} & \begin{tabular}{@{}c@{}}$\lambda_1 = 0.06$, $\lambda_2 = 0.01$ \\  dim = 12, $\alpha = 8$\end{tabular}\\
$S^3C$ & $\gamma = 1$, $\alpha = 20$ & $\gamma = 1$, $\alpha = 100$ &  $\gamma = 1$, $\alpha = 20$\\
LRSC & $\tau = 0.045$,$\alpha=10^5$ & $\tau = 0.07$,$\alpha=0.1$ & $\tau = 0.045$,$\alpha=0.07$ \\

\hline
\end{tabular} 
\end{center}
\end{table*}
\end{center}

 \begin{center}
\begin{table*}[!htbp]
\begin{center}
\caption{Average performance on the Extended Yale B data set with different number of subjects. The best performance is indicated in bold.}
\label{tabyale}  
\begin{tabular}{|cccccccc|}
\hline
Algorithm	& \multicolumn{1}{|c|}{LG-SSC} & \multicolumn{1}{|c|}{MG-SSC} & \multicolumn{1}{|c|}{SSC} & \multicolumn{1}{|c|}{LRR} & \multicolumn{1}{|c|}{EDSC} & \multicolumn{1}{|c|}{$S^3C$} &\multicolumn{1}{|c|}{LRSC} \\
\hline
 \multicolumn{1}{c}{2 subjects} & &&&& \multicolumn{1}{c}{}\\
\hline
ACC & \multicolumn{1}{|c|}{\textbf{99.92}} & \multicolumn{1}{|c|}{99.91} & \multicolumn{1}{|c|}{98.14} & \multicolumn{1}{|c|}{89.69} & \multicolumn{1}{|c|}{97.35$^*$} &\multicolumn{1}{|c|}{99.48$^*$} &\multicolumn{1}{|c|}{96.23} \\
NMI & \multicolumn{1}{|c|}{\textbf{99.54}} & \multicolumn{1}{|c|}{99.38} & \multicolumn{1}{|c|}{93.16} & \multicolumn{1}{|c|}{66.69} & \multicolumn{1}{|c|}{-} &\multicolumn{1}{|c|}{-} &\multicolumn{1}{|c|}{82.05}\\
ARI & \multicolumn{1}{|c|}{\textbf{99.70}} & \multicolumn{1}{|c|}{99.66} & \multicolumn{1}{|c|}{94.29} & \multicolumn{1}{|c|}{68.13} & \multicolumn{1}{|c|}{-} &\multicolumn{1}{|c|}{-}&\multicolumn{1}{|c|}{86.23}\\
\hline
 \multicolumn{1}{c}{3 subjects} & &&&& \multicolumn{1}{c}{}\\
\hline
ACC & \multicolumn{1}{|c|}{99.42} & \multicolumn{1}{|c|}{\textbf{99.87}} & \multicolumn{1}{|c|}{96.70} & \multicolumn{1}{|c|}{79.09} & \multicolumn{1}{|c|}{96.35$^*$} &\multicolumn{1}{|c|}{99.11$^*$} &\multicolumn{1}{|c|}{93.55} \\
NMI & \multicolumn{1}{|c|}{99.04} & \multicolumn{1}{|c|}{\textbf{99.43}} & \multicolumn{1}{|c|}{92.75} & \multicolumn{1}{|c|}{59.61} & \multicolumn{1}{|c|}{-} &\multicolumn{1}{|c|}{-} &\multicolumn{1}{|c|}{81.06}\\
ARI & \multicolumn{1}{|c|}{98.99} & \multicolumn{1}{|c|}{\textbf{99.62}} & \multicolumn{1}{|c|}{92.61} & \multicolumn{1}{|c|}{53.22} & \multicolumn{1}{|c|}{-} &\multicolumn{1}{|c|}{-}&\multicolumn{1}{|c|}{82.61}\\
\hline
 \multicolumn{1}{c}{5 subjects} & &&&& \multicolumn{1}{c}{}\\
\hline
ACC & \multicolumn{1}{|c|}{99.35} & \multicolumn{1}{|c|}{\textbf{99.78}} & \multicolumn{1}{|c|}{95.68} & \multicolumn{1}{|c|}{65.46} & \multicolumn{1}{|c|}{94.89$^*$} &\multicolumn{1}{|c|}{98.49$^*$} &\multicolumn{1}{|c|}{90.46} \\
NMI & \multicolumn{1}{|c|}{99.02} & \multicolumn{1}{|c|}{\textbf{99.32}} & \multicolumn{1}{|c|}{91.56} & \multicolumn{1}{|c|}{54.53} & \multicolumn{1}{|c|}{-} &\multicolumn{1}{|c|}{-} &\multicolumn{1}{|c|}{80.74}\\
ARI & \multicolumn{1}{|c|}{98.85} & \multicolumn{1}{|c|}{\textbf{99.46}} & \multicolumn{1}{|c|}{90.17} & \multicolumn{1}{|c|}{39.07} & \multicolumn{1}{|c|}{-} &\multicolumn{1}{|c|}{-}&\multicolumn{1}{|c|}{78.99}\\
\hline
 \multicolumn{1}{c}{8 subjects} & &&&& \multicolumn{1}{c}{}\\
\hline
ACC & \multicolumn{1}{|c|}{99.41} & \multicolumn{1}{|c|}{\textbf{99.72}} & \multicolumn{1}{|c|}{94.13} & \multicolumn{1}{|c|}{59.02} & \multicolumn{1}{|c|}{93.93$^*$} &\multicolumn{1}{|c|}{97.69$^*$} &\multicolumn{1}{|c|}{76.36} \\
NMI & \multicolumn{1}{|c|}{98.54} & \multicolumn{1}{|c|}{\textbf{99.35}} & \multicolumn{1}{|c|}{90.58} & \multicolumn{1}{|c|}{56.34} & \multicolumn{1}{|c|}{-} &\multicolumn{1}{|c|}{-} &\multicolumn{1}{|c|}{70.71}\\
ARI & \multicolumn{1}{|c|}{98.65} & \multicolumn{1}{|c|}{\textbf{99.38}} & \multicolumn{1}{|c|}{86.44} & \multicolumn{1}{|c|}{36.27} & \multicolumn{1}{|c|}{-} &\multicolumn{1}{|c|}{-}&\multicolumn{1}{|c|}{59.15}\\
\hline
 \multicolumn{1}{c}{10 subjects} & &&&& \multicolumn{1}{c}{}\\
\hline
ACC & \multicolumn{1}{|c|}{\textbf{99.68}} & \multicolumn{1}{|c|}{\textbf{99.68}} & \multicolumn{1}{|c|}{92.60} & \multicolumn{1}{|c|}{60.42} & \multicolumn{1}{|c|}{92.76$^*$} &\multicolumn{1}{|c|}{97.19$^*$} &\multicolumn{1}{|c|}{66.56} \\
NMI & \multicolumn{1}{|c|}{\textbf{99.33}} & \multicolumn{1}{|c|}{\textbf{99.33}} & \multicolumn{1}{|c|}{89.37} & \multicolumn{1}{|c|}{59.79} & \multicolumn{1}{|c|}{-} &\multicolumn{1}{|c|}{-} &\multicolumn{1}{|c|}{66.27}\\
ARI & \multicolumn{1}{|c|}{\textbf{99.31}} & \multicolumn{1}{|c|}{\textbf{99.31}} & \multicolumn{1}{|c|}{82.72} & \multicolumn{1}{|c|}{38.13} & \multicolumn{1}{|c|}{-} &\multicolumn{1}{|c|}{-}&\multicolumn{1}{|c|}{49.23}\\
\hline
\end{tabular} 
\end{center}
\end{table*}
\end{center}

 \begin{center}
\begin{table*}[!htbp]
\begin{center}
\caption{Performance on the Extended Yale B data set with different number of subjects. The best performance is indicated in bold.}
\label{tabyale2}  
\begin{tabular}{c|c||ccccccc}
\hline
\#subjects & Metric & LG-SSC & MG-SSC & SSC & LRR & EDSC & S$^3$C & LRSC \\
\hline
15 & \begin{tabular}{@{}c@{}c@{}}ACC \\ NMI \\ ARI\end{tabular} & \begin{tabular}{@{}c@{}c@{}}99.47 \\ 99.10 \\ 98.87\end{tabular}&\begin{tabular}{@{}c@{}c@{}} \textbf{100} \\ \textbf{100} \\ \textbf{100}\end{tabular}&\begin{tabular}{@{}c@{}c@{}}78.81 \\ 79.14 \\ 60.89\end{tabular} &\begin{tabular}{@{}c@{}c@{}}64.30 \\ 66.18 \\ 41.18\end{tabular} &\begin{tabular}{@{}c@{}c@{}}86.44 \\ 88.97 \\ 80.13 \end{tabular} & \begin{tabular}{@{}c@{}c@{}}88.24 \\ 91.25 \\ 84.94 \end{tabular} & \begin{tabular}{@{}c@{}c@{}}68.75 \\ 71.57 \\ 51.65\end{tabular}\\
\hline
20 & \begin{tabular}{@{}c@{}c@{}}ACC \\ NMI \\ ARI\end{tabular} & \begin{tabular}{@{}c@{}c@{}}\textbf{98.73} \\ \textbf{98.05} \\ \textbf{97.31}\end{tabular}&\begin{tabular}{@{}c@{}c@{}} 98.65\\ 98.02 \\ 97.04\end{tabular}&\begin{tabular}{@{}c@{}c@{}}73.61 \\ 76.67 \\ 54.50\end{tabular} &\begin{tabular}{@{}c@{}c@{}}68.07 \\ 70.68 \\ 42.99 \end{tabular} &\begin{tabular}{@{}c@{}c@{}}88.51 \\ 90.79 \\ 81.98 \end{tabular} & \begin{tabular}{@{}c@{}c@{}}85.73 \\ 91.15 \\ 82.79 \end{tabular} & \begin{tabular}{@{}c@{}c@{}} 71.08 \\ 75.49 \\ 52.90\end{tabular}\\
\hline
30 & \begin{tabular}{@{}c@{}c@{}}ACC \\ NMI \\ ARI\end{tabular} & \begin{tabular}{@{}c@{}c@{}}\textbf{98.69} \\ \textbf{98.09} \\ \textbf{97.27}\end{tabular}&\begin{tabular}{@{}c@{}c@{}} 92.43\\ 94.57 \\ 88.35\end{tabular}&\begin{tabular}{@{}c@{}c@{}}74.66 \\ 77.69 \\ 51.20\end{tabular} &\begin{tabular}{@{}c@{}c@{}} 71.50 \\ 75.35 \\ 43.90\end{tabular} &\begin{tabular}{@{}c@{}c@{}}87.22 \\ 91.22 \\ 79.46 \end{tabular} & \begin{tabular}{@{}c@{}c@{}} 84.91\\ 90.48 \\ 80.63 \end{tabular} & \begin{tabular}{@{}c@{}c@{}} 71.24 \\ 75.19 \\ 52.90\end{tabular}\\
\hline
38 & \begin{tabular}{@{}c@{}c@{}}ACC \\ NMI \\ ARI\end{tabular} & \begin{tabular}{@{}c@{}c@{}}\textbf{93.37}\\\textbf{ 94.91} \\ \textbf{86.03}\end{tabular}&\begin{tabular}{@{}c@{}c@{}} 90.27\\ 91.47 \\ 78.99\end{tabular}&\begin{tabular}{@{}c@{}c@{}}70.67 \\ 75.44 \\ 40.52\end{tabular} &\begin{tabular}{@{}c@{}c@{}}66.28 \\ 72.19 \\ 45.99\end{tabular} &\begin{tabular}{@{}c@{}c@{}}85.29 \\ 90.08 \\ 72.67 \end{tabular} & \begin{tabular}{@{}c@{}c@{}}78.71 \\ 86.78 \\ 68.16 \end{tabular} & \begin{tabular}{@{}c@{}c@{}}70.17 \\ 75.19 \\ 52.46\end{tabular}\\
\hline
\end{tabular} 
\end{center}
\end{table*}
\end{center}

\subsection{AR face data set}
The AR database~\cite{martinez1998ar} contains frontal face images for 100 individuals (50 men and 50 females). There are 26 colored pictures collected for each person. The images include facial variations such as illumination changes, different expressions and facial disguises using sunglasses and scarves. Compared to Extended Yale B, this database is more challenging because of occlusions and fewer number of images per individual. We downsampled each image to $48 \times 42$ pixels and converted them to gray scale. We set $p=4$ and $s=3$. The performance of LG-SSC with respect to different number of clusters, is compared with MG-SSC, SSC, LRR, LRSC, EDSC and $S^3C$ in table~\ref{tabar}. We observe that:

\begin{itemize}
\item Performance of almost all approaches (except LRR, MG-SSC and LG-SSC) degraded compared to Extended Yale B database. This result is expected as AR database is a more challenging database and with higher number of clusters.
\item LG-SSC has a better performance compared to other approaches in all cases by a large margin. The robustness of LG-SSC is clearly evident for this database and the patch-based representations are elegantly guiding the global self-expressive representation to a more robust clustering segmentation. 
\item LG-SSC consistantly performs better than MG-SSC which further highlights the efficiency of LG-SSC in combining the information of local patches with calculation of robust global self-expressive representation.
\item The occlusions are degrading the performance of SSC, EDSC, $S^3C$ and LRSC even in the simplest case of 5 clusters. This shows the sensitivity of these approaches to the occlusions and contiguous corruptions.
\item LRSC attempts to recover a clean dictionary by optimizing a nonconvex problem, however this approach assumes that the data is contaminated by sparse error which is clearly violated in this database.
\item The post-processing step of EDSC cannot improve the performance in this case. This is due to the severely corrupted global representation that makes the \emph{correction} difficult (if not impossible).
\item The third best performance is achieved by LRR. The LRR approach is considered as the extension of RPCA~\cite{candes2011robust} for the union of subspaces. Dense representations of nuclear norm appears to be more suitable compared to sparse representations for the data with complex noise structures.
\end{itemize}

\begin{figure*}[htb]
\begin{minipage}[b]{0.3\linewidth}
  \centering
  \centerline{\includegraphics[width=6cm]{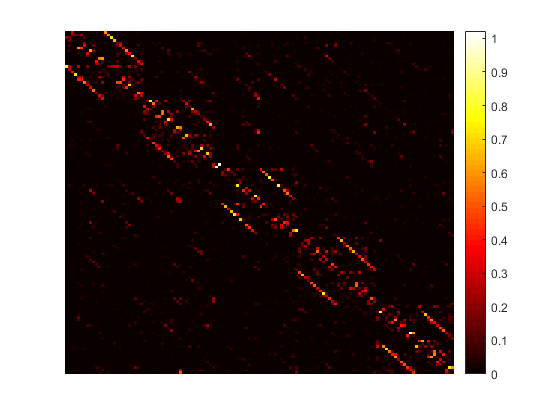}}
  \centerline{(a) SSC}\medskip
\end{minipage}
\hfill
\begin{minipage}[b]{0.3\linewidth}
  \centering
  \centerline{\includegraphics[width=6cm]{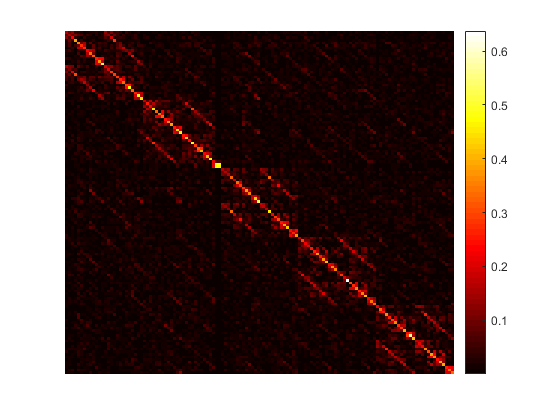}}
  \centerline{(b) LRR}\medskip
\end{minipage}
\hfill
\begin{minipage}[b]{0.3\linewidth}
  \centering
  \centerline{\includegraphics[width=6cm]{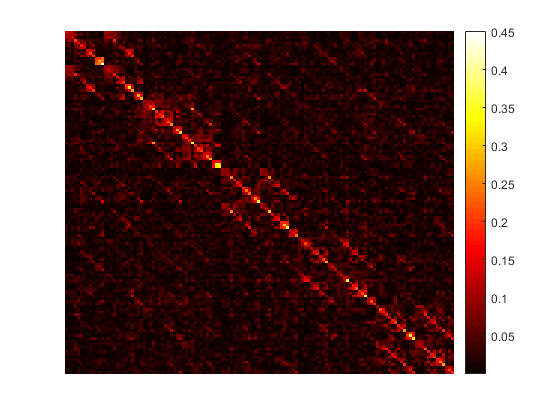}}
  \centerline{(c) EDSC}\medskip
\end{minipage}
\hfill
\begin{minipage}[b]{0.3\linewidth}
  \centering
  \centerline{\includegraphics[width=6cm]{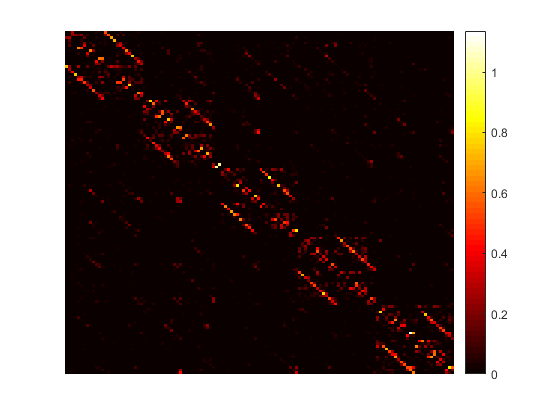}}
  \centerline{(d) $S^3C$}\medskip
\end{minipage}
\hfill
\begin{minipage}[b]{0.3\linewidth}
  \centering
  \centerline{\includegraphics[width=6cm]{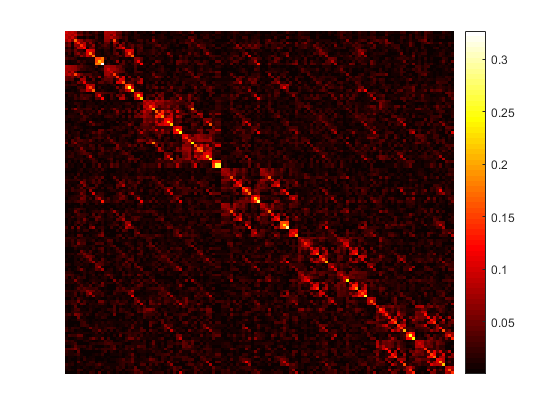}}
  \centerline{(e) LRSC}\medskip
\end{minipage}
\hfill
\begin{minipage}[b]{0.3\linewidth}
  \centering
  \centerline{\includegraphics[width=6cm]{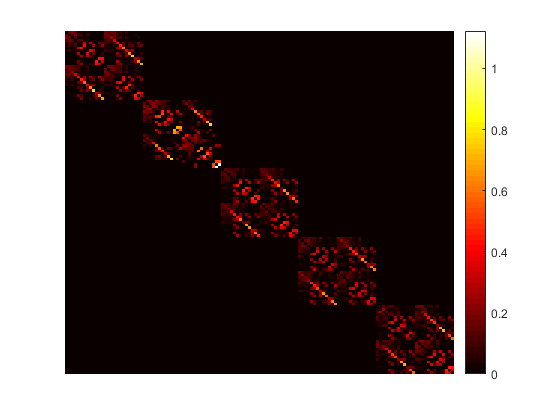}}
  \centerline{(f) LG-SSC}\medskip
\end{minipage}
\caption{Coefficient matrices of (a) SSC, (b) LRR, (c) EDSC, (d) $S^3C$, (e) LRSC and (f) LG-SSC for the first 5 individuals of AR database.}
\label{AR_C}
\end{figure*}

For better visualization and comparison, the coefficient matrices corresponding to each algorithm for the first 5 individuals are plotted in Figure~\ref{AR_C}. The block diagonal structure in  LG-SSC's coefficient matrix is clearly evident. Two major components for the success of a subspace clustering algorithm, namely, (i) subspace preserving connections and (ii) strong connectivity withing each subspace is present for LG-SSC. However, the coefficient matrices of other approaches are contaminated due to illumination effects and disguises which is affecting the clustering performance as well. Note that MG-SSC does not output a final coefficient matrix, hence the comparison is done with other 5 approaches.

 \begin{center}
\begin{table*}[!htbp]
\begin{center}
\caption{Performance on the AR data set with different number of subjects. The best accuracy is indicated in bold.}
\label{tabar}  
\begin{tabular}{c|c||ccccccc}
\hline
\#subjects & Metric & LG-SSC & MG-SSC & SSC & LRR & EDSC & S$^3$C & LRSC \\
\hline
5 & \begin{tabular}{@{}c@{}c@{}}ACC \\ NMI \\ ARI\end{tabular} & \begin{tabular}{@{}c@{}c@{}}\textbf{100} \\ \textbf{100} \\ \textbf{100}\end{tabular}&\begin{tabular}{@{}c@{}c@{}} \textbf{100} \\ \textbf{100} \\\textbf{100} \end{tabular}&\begin{tabular}{@{}c@{}c@{}}76.92 \\ 62.79 \\ 48.32\end{tabular} &\begin{tabular}{@{}c@{}c@{}}82.31 \\ 71.57 \\ 62.33\end{tabular} &\begin{tabular}{@{}c@{}c@{}} 75.38 \\ 64.74 \\ 56.17 \end{tabular} & \begin{tabular}{@{}c@{}c@{}}76.92 \\ 64.04 \\ 48.55 \end{tabular} & \begin{tabular}{@{}c@{}c@{}}63.08 \\ 47.87 \\ 35.55\end{tabular}\\
\hline
10 & \begin{tabular}{@{}c@{}c@{}}ACC \\ NMI \\ ARI\end{tabular} & \begin{tabular}{@{}c@{}c@{}}\textbf{100} \\ \textbf{100} \\ \textbf{100}\end{tabular}&\begin{tabular}{@{}c@{}c@{}} \textbf{100} \\ \textbf{100} \\\textbf{100} \end{tabular}&\begin{tabular}{@{}c@{}c@{}}66.54 \\ 65.34 \\ 41.97\end{tabular} &\begin{tabular}{@{}c@{}c@{}}68.08 \\ 72.27 \\ 55.13\end{tabular} &\begin{tabular}{@{}c@{}c@{}} 73.46 \\ 81.29 \\ 65.89 \end{tabular} & \begin{tabular}{@{}c@{}c@{}}71.15\\ 73.19 \\ 50.13 \end{tabular} & \begin{tabular}{@{}c@{}c@{}}67.69 \\ 68.64 \\ 52.11\end{tabular}\\
\hline
20 & \begin{tabular}{@{}c@{}c@{}}ACC \\ NMI \\ ARI\end{tabular} & \begin{tabular}{@{}c@{}c@{}}\textbf{100} \\ \textbf{100} \\ \textbf{100}\end{tabular}&\begin{tabular}{@{}c@{}c@{}} 90.00 \\ 92.82 \\86.38 \end{tabular}&\begin{tabular}{@{}c@{}c@{}}59.42 \\ 68.86 \\ 40.41\end{tabular} &\begin{tabular}{@{}c@{}c@{}}80.58 \\ 86.14 \\ 73.89\end{tabular} &\begin{tabular}{@{}c@{}c@{}} 66.16 \\ 75.61 \\ 51.78 \end{tabular} & \begin{tabular}{@{}c@{}c@{}}60.00\\ 69.25 \\ 37.84 \end{tabular} & \begin{tabular}{@{}c@{}c@{}}72.69 \\ 77.42 \\ 55.68\end{tabular}\\
\hline
50 & \begin{tabular}{@{}c@{}c@{}}ACC \\ NMI \\ ARI\end{tabular} & \begin{tabular}{@{}c@{}c@{}}\textbf{96.15} \\ \textbf{97.71} \\ \textbf{94.28}\end{tabular}&\begin{tabular}{@{}c@{}c@{}} 86.15 \\ 90.33 \\ 75.09 \end{tabular}&\begin{tabular}{@{}c@{}c@{}}67.31 \\ 78.21 \\ 47.21\end{tabular} &\begin{tabular}{@{}c@{}c@{}}87.31 \\ 91.79 \\ 75.85\end{tabular} &\begin{tabular}{@{}c@{}c@{}} 65.76 \\ 79.95 \\ 51.56 \end{tabular} & \begin{tabular}{@{}c@{}c@{}} 58.08\\ 72.90 \\ 31.95 \end{tabular} & \begin{tabular}{@{}c@{}c@{}} 69.15 \\ 79.76 \\ 56.30\end{tabular}\\
\hline
75 & \begin{tabular}{@{}c@{}c@{}}ACC \\ NMI \\ ARI\end{tabular} & \begin{tabular}{@{}c@{}c@{}}\textbf{94.97} \\ \textbf{96.93} \\ \textbf{92.69}\end{tabular}&\begin{tabular}{@{}c@{}c@{}} 87.08 \\ 91.23 \\ 70.93 \end{tabular}&\begin{tabular}{@{}c@{}c@{}}67.64 \\ 82.11 \\ 53.34\end{tabular} &\begin{tabular}{@{}c@{}c@{}}84.26 \\ 91.39 \\ 75.85\end{tabular} &\begin{tabular}{@{}c@{}c@{}} 67.69 \\ 83.69 \\ 55.32 \end{tabular} & \begin{tabular}{@{}c@{}c@{}} 61.49\\ 78.17 \\ 38.81 \end{tabular} & \begin{tabular}{@{}c@{}c@{}} 69.49 \\ 81.55 \\ 59.13 \end{tabular}\\
\hline
100 & \begin{tabular}{@{}c@{}c@{}}ACC \\ NMI \\ ARI\end{tabular} & \begin{tabular}{@{}c@{}c@{}}\textbf{90.00} \\ \textbf{93.74} \\ \textbf{83.06}\end{tabular}&\begin{tabular}{@{}c@{}c@{}} 83.27 \\ 90.98 \\ 73.02 \end{tabular}&\begin{tabular}{@{}c@{}c@{}}68.15 \\ 82.87 \\ 52.52\end{tabular} &\begin{tabular}{@{}c@{}c@{}}79.92 \\ 90.01 \\ 71.06\end{tabular} &\begin{tabular}{@{}c@{}c@{}} 67.54 \\ 82.77 \\ 52.29 \end{tabular} & \begin{tabular}{@{}c@{}c@{}} 60.54\\ 79.89 \\ 43.09 \end{tabular} & \begin{tabular}{@{}c@{}c@{}} 67.23 \\ 82.34 \\ 57.25 \end{tabular}\\
\hline
\end{tabular} 
\end{center}
\end{table*}
\end{center}

\subsection{Coil-20 data set}
Columbia Object Image Library(COIL-20)~\cite{nene1996columbia} contains 1440 gray-scale images of 20 objects in different poses. The images of the objects were taken by placing objects on a turntable against a black background. There are 72 images per object, each of size $128\times128$ pixels. We downsampled each image to $32 \times 32$ pixels. Even though, this database is clean, with no noise or occlusions, it would be still interesting to observe the performance of LG-SSC in dealing with data from clean subspaces. We divided each image to 9 overlapping patches and set $s=2$. This is due to the fact that for images of the objects, the patches closer to the center of image contain more meaningful information compared to other patches.The performance of our proposed LG-SSC is compared with other subspace clustering algorithms in Table~\ref{tabcoil}. We can conclude that:
\begin{itemize}
\item Even though MG-SSC fails to increase the accuracy in dealing with clean images of objects, LG-SSC is able to increase the clustering accuracy by more than 10\%. This suggests that locally guided self-expressiveness might improve the quality of clustering in the challenging case of close subspaces.
\item EDSC enjoys the benefits of post-processing the coefficient matrix and has the second best performance. 
\item sparse based approaches (SSC and $S^3C$) have higher performance compared to low-rank based algorithms (LRR and LRSC). In general sparse based approaches have stronger theoretical guarantees compared to low-rank based alternatives and hence, for the clean data sets, they are usually expected to perform better. 

\end{itemize}

 \begin{center}
\begin{table*}[!htbp]
\begin{center}
\caption{Performance on the COIL-20 data set with different number of clusters. The best accuracy is indicated in bold.}
\label{tabcoil}  
\begin{tabular}{c||ccccccc}
\hline
Metric & LG-SSC & MG-SSC & SSC & LRR & EDSC & S$^3$C & LRSC \\
\hline
\begin{tabular}{@{}c@{}c@{}}ACC \\ NMI \\ ARI\end{tabular} & \begin{tabular}{@{}c@{}c@{}}\textbf{89.58} \\ \textbf{95.34} \\ \textbf{85.53}\end{tabular}&\begin{tabular}{@{}c@{}c@{}} 78.26 \\ 87.92 \\ 72.30\end{tabular}&\begin{tabular}{@{}c@{}c@{}}78.68 \\ 90.39 \\74.89 \end{tabular} &\begin{tabular}{@{}c@{}c@{}}54.86 \\ 70.03 \\ 42.19 \end{tabular} &\begin{tabular}{@{}c@{}c@{}} 84.51 \\ 93.52 \\ 81.54 \end{tabular} & \begin{tabular}{@{}c@{}c@{}}74.86 \\ 88.28 \\ 66.88 \end{tabular} & \begin{tabular}{@{}c@{}c@{}}64.09 \\ 72.29 \\ 52.22\end{tabular}\\
\hline
\end{tabular} 
\end{center}
\end{table*}
\end{center}

\subsection{Parameter Analysis}
In LG-SSC, there are three parameters that are used in the optimization problem~(\ref{final}) which controls the trade-off between four qualities: (i) sparsity, (ii) ignoring cannot-links, (iii) respecting the recommended-links and (iv) the self-expressive reconstruction error. Following the methodology in~\cite{elhamifar2013sparse}, we set the regularization parameter $\mu$ as $\frac{\alpha}{\max_{j\neq i} |x_i^Tx_j|}$ where $\alpha >1$ is tuned for each dataset. In this paper, we set this parameter to the values that were commonly used and reported in subspace clustering literature. 

\begin{figure*}[htb]
\begin{minipage}[b]{0.3\linewidth}
  \centering
  \centerline{\includegraphics[width=6cm]{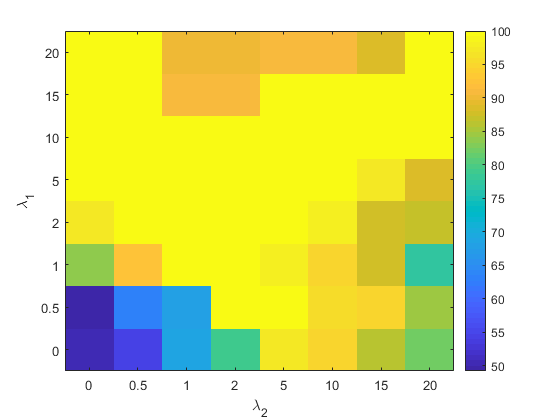}}
  \centerline{(a) AR database}\medskip
\end{minipage}
\hfill
\begin{minipage}[b]{0.3\linewidth}
  \centering
  \centerline{\includegraphics[width=6cm]{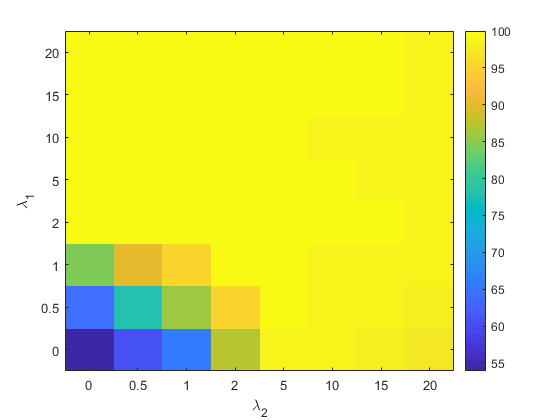}}
  \centerline{(b) YALE B database}\medskip
\end{minipage}
\hfill
\begin{minipage}[b]{0.3\linewidth}
  \centering
  \centerline{\includegraphics[width=6cm]{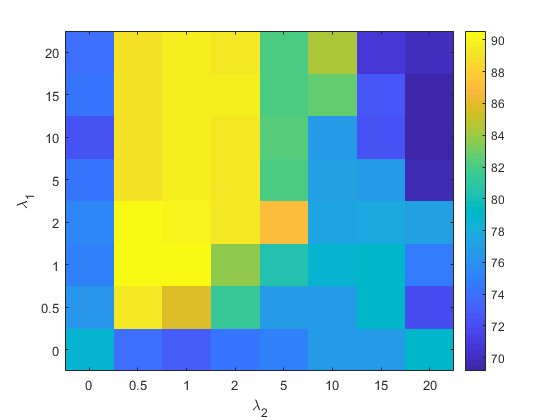}}
  \centerline{(c) COIL-20 database}\medskip
\end{minipage}
\caption{Effect of $\lambda_1$ and $\lambda_2$ on the accuracy of LG-SSC for (a) AR database, (b) Extended Yale B database and (c) COIL-20 database}
\label{param}
\end{figure*}

The behavior of LG-SSC with respect to $\lambda_1$ and $\lambda_2$ is empirically validated on all three databases (AR, Extended Yale B and Coil-20). We consider the clustering performance for the first 10 subjects of AR and Yale B and 20 objects of Coil-20. The clustering accuracy with respect to different values of these parameters is illustrated for each database in Figure~\ref{param}. It can be seen that for values of $\lambda_1 \in [2:10]$ and $\lambda_2 \in [0.5:2]$, the accuracy is quite stable for all three cases. In particular, Yale B is the least sensitive one to the values of $\lambda_1$ and $\lambda_2$ and Coil-20 is the most sensitive database. For Yale B database, as long as $\lambda_1$ and $\lambda_2$ are not too small, the accuracy is almost 100\%. Interestingly by setting the $\lambda_1 \in [2:20]$ and $\lambda_2 = 0$, the accuracy is still 100\%. This suggests that for this database, the "cannot-links" information is more important compared to the "recommended-links" information. However, for Coil-20, the recommended-links information plays an important role in boosting the accuracy of the basic SSC (which is around 78.68\%).

We also evaluate the effect of patch sizes ($p$) and the number of levels ($s$) on the clustering accuracy. We consider $s \in \{2,3,4\}$ and $p \in \{2,3,4\}$. The performance of LG-SSC for different values of $s$ and $p$ for the first 10 subjects of Yale B and AR and 20 objects of Coil-20 are reported in Table~\ref{tabblocklevel}. For the AR database, the clustering accuracy is not affected by the patch-size as long as $s=3$. Because for $s=2, p=2$ and $s=2, p =3$, the patches at the coarse level are not robust themselves and they contain occluded parts of image, hence, the robustness is not transferred to the fine scale. This can be confirmed by considering the case where $s=2$ and $p=4$. In this case, the accuracy is 100\% because the patches in the second level are small enough to contain robust discriminant information. By increasing the number of levels and patches to 4 ($s=4$ and $p=4$), the accuracy decreases significantly to 22.31\%. In this case, the patches at the last level are very small and hence, neither robust nor discriminant information can be fed into upper levels. For Yale B, which is relatively less challenging compared to AR, the accuracy is almost 100\% in all cases except for the case with $s=4$ and $p=4$ where the patches get intuitively very small. For this database, $s=2$ is sufficient to increase robustness to illumination variations. Interestingly, the best accuracy for the Coil-20 database is achieved only for $s=2$ and $p=2$. Note that in this case, each image is $32 \times 32$ pixels and hence we do not consider $s=4$. For object clustering, the edges play a critical role, hence the patches should be considered such that they contain enough edge information for accurate clustering.

\begin{center}
\begin{table*}[!htbp]
\begin{center}
\caption{Accuracy of LG-SSC with respect to different values for levels ($s$) and number of blocks in each level ($p$)}
\label{tabblocklevel}  
\small\addtolength{\tabcolsep}{-1pt}
\begin{tabular}{c||ccc|ccc|ccc}
\hline
	 \multicolumn{1}{c||}{} & \multicolumn{3}{c|}{$2 \times 2$} & \multicolumn{3}{c|}{$3 \times 3$}& \multicolumn{3}{c}{$4 \times 4$}  \\
	  & s=2 & s=3& s=4 & s=2 & s=3 & s=4 & s=2 & s=3& s=4\\
\hline
 AR (10) & 60.38 & 100 & 99.61 & 86.54 & 100 & 99.62 & 100 & 100 & 22.31\\
YALE B (10) & 100 & 100 & 99.84 & 99.84 & 100 & 99.68 & 99.68 & 100 & 18.13\\
 COIL-20 & 89.44 & 77.84 & - & 77.22 & 74.16  & - &74.37  &  72.5 & -\\
\hline
\end{tabular} 
\end{center}
\end{table*}
\end{center}

\subsection{Neither Global Nor Local}
In this section, we discuss the role of multi-layer graph fusion approach and emphasize on the point that almost neither of individual local patches nor global data might lead to a robust discriminant representation. However, merging the local representations using their low-dimensional embedding on Grassmann manifold can provide a summary representation which highlights the information that majority of local representations tend to agree on. Hence, the clustering accuracy of each local patch for the three databases (all samples for each database are considered) are plotted in Figure~\ref{local}. For the AR dataset, two cases of $p=4$ and $p=16$ are considered. As can be seen, none of local patches reach an accuracy higher than 65\% but applying a k-means on summarized low-dimensional embedding of these 16 patches lead to an accuracy near 85\% (first column from right). LG-SSC further boosts this robustness to near 90\%. When $p=4$, the same observation in Table~\ref{tabblocklevel} is repeated and not only neither of local patches have an accuracy higher than 65\% but also the merged information of these 4 local patches does not boost the performance significantly. As mentioned previously, this is because none of these patches have robust representations. In Yale B, the coefficient matrix corresponding to the 4th patch has the highest clustering accuracy and LG-SSC is improving this accuracy without getting affected by other patches, eg.  the 1st patch.
For Coil-20 database, not only the local patches do not lead to high clustering accuracy but also the merged low-dimensional embedding does not increase the clustering accuracy significantly as well. However, LG-SSC still increases the clustering accuracy. This is due to the fact that the merged information \emph{induces} the global sparse self-expressive representation and for this database, the local representations provide sufficient information to avoid miss-clustering of closely related objects.

\begin{figure*}[htb]
\begin{minipage}[b]{0.2\linewidth}
  \centering
  \centerline{\includegraphics[width=5cm]{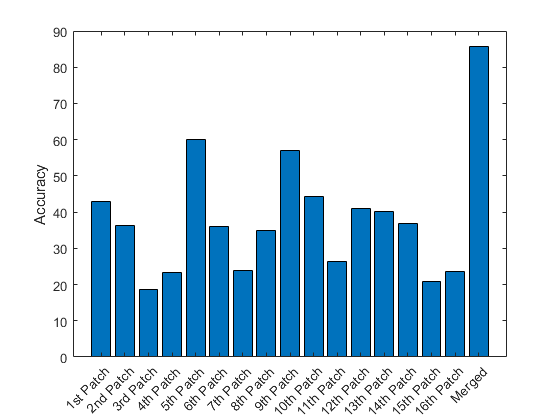}}
  \centerline{(a) AR (16 patches)}\medskip
\end{minipage}
\hfill
\begin{minipage}[b]{0.2\linewidth}
  \centering
  \centerline{\includegraphics[width=5cm]{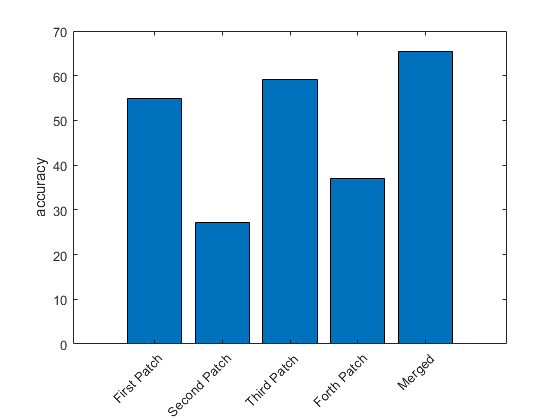}}
  \centerline{(b) AR (4 patches)}\medskip
\end{minipage}
\hfill
\begin{minipage}[b]{0.2\linewidth}
  \centering
  \centerline{\includegraphics[width=5cm]{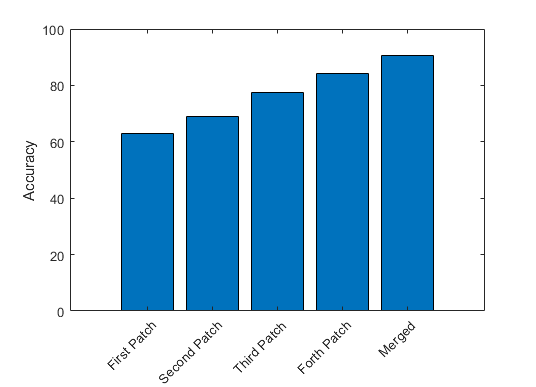}}
  \centerline{(c) Yale B database}\medskip
\end{minipage}
\hfill
\begin{minipage}[b]{0.2\linewidth}
  \centering
  \centerline{\includegraphics[width=5cm]{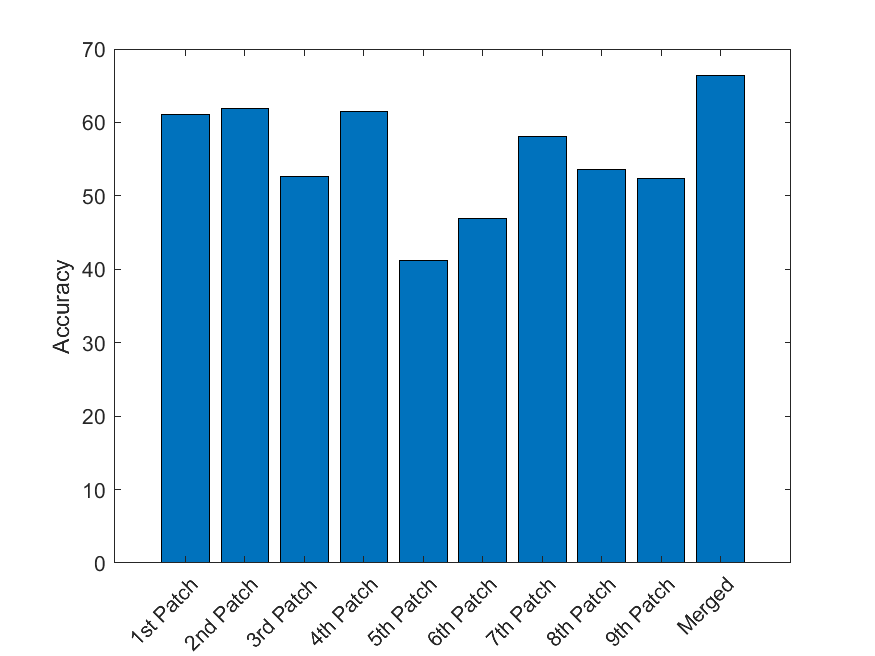}}
  \centerline{(d) COIL-20 database}\medskip
\end{minipage}
\caption{Clustering accuracy of individual local patches and the corresponding merged low-dimensional embedding on (a) AR (dividing to 16 patches), (b) AR (dividing to 4 patches), (c) Extended Yale B and (d) Coil-20.}
\label{local2}
\end{figure*}

\section {Conclusion} \label{conclusion}
In this paper, we uncovered the importance of local representations in improving the robustness of self-expressive based subspace clustering approaches. The proposed hierarchical approach bridges the gap between robust local representations and discriminant global alternative in order to obtain a robust discriminant self-expressive representation for the input data. This approach consists of two major key ingredients: 1) Efficiently summarizing local based representations using low-rank embedding on a Grassmann manifold to obtain cannot-links and recommended links which local patches agree on them. 2) Employing this summarized information into the optimization problem for calculating self-expressive representation in each level using weighted group lasso regularization. Robustness of proposed approach to occlusion and complex noise was confirmed by experimental results.

\small 

\bibliographystyle{spmpsci} 
\bibliography{LGSSC}

\end{document}